\documentclass{article}
\usepackage[T1]{fontenc}
\usepackage{spconf,amsmath,graphicx}

\usepackage{enumitem}
\setlist{nosep, leftmargin=14pt}

\makeatletter
\renewcommand\section{\@startsection {section}{1}{\z@}%
                                   {-2.5ex \@plus -1ex \@minus -.2ex}%
                                   {1.3ex \@plus.2ex}%
                                   {\normalfont\Large\bfseries}}
\makeatother

\usepackage{amssymb,amsfonts}
\usepackage{algorithmic}
\usepackage{graphicx}
\usepackage{textcomp}
\usepackage{booktabs}
\usepackage{xcolor}

\usepackage{url}
\usepackage[square,sort&compress,comma,numbers]{natbib}
\usepackage{setspace}
\usepackage{fancyhdr}
\newcommand*{\doi}[1]{{\scriptsize doi:\href{http://dx.doi.org/\detokenize{#1}}{\detokenize{#1}}}}

\usepackage[colorlinks=true]{hyperref}

\usepackage[capitalize]{cleveref}

\usepackage{mwe} %

\clearpage{}%
\usepackage{xspace}

\makeatletter
\DeclareRobustCommand\onedot{\futurelet\@let@token\@onedot}
\def\@onedot{\ifx\@let@token.\else.\null\fi\xspace}

\def\eg{\emph{e.g}\onedot} 
\def\ie{\emph{i.e}\onedot} 
\def\cf{\emph{cf.}\xspace}

\def\wrt{w.r.t\onedot} 

\makeatother

\newcommand{\stardist}{\mbox{\small\textsc{StarDist}}\xspace}
\newcommand{\stardistfns}{\mbox{\footnotesize\textsc{StarDist}}\xspace}

\newcommand{\PQ}{\mathrm{PQ}}
\newcommand{\DQ}{\mathrm{DQ}}
\newcommand{\SQ}{\mathrm{SQ}}
\newcommand{\mPQ}{\mathrm{mPQ}}\clearpage{}%
\title{Nuclei instance segmentation and classification in histopathology images with StarDist}
\name{Martin Weigert$^1$, Uwe Schmidt$^2$ }

\address{
  $^1$Institute of Bioengineering, School of Life Sciences, EPFL, Switzerland\\
  $^2$Independent Researcher, Dresden, Germany}

\begin{document}
\maketitle
\thispagestyle{fancy}

\begin{abstract}
  Instance segmentation and classification of nuclei is an important task in
  computational pathology. We show that \emph{StarDist}, a deep learning
  nuclei segmentation method originally developed for fluorescence microscopy,
  can be extended and successfully applied to histopathology images. This is
  substantiated by conducting experiments on the \emph{Lizard} dataset, and
  through entering the \emph{Colon Nuclei Identification and Counting (CoNIC)}
  challenge 2022, where our approach achieved the first spot on the leaderboard for the segmentation and classification task for both the preliminary and final test phase.

\end{abstract}

\section{Introduction}
\label{sec:intro}

Reliably identifying individual cell nuclei in microscopy images is a ubiquitous
task in the life sciences. This can be especially challenging when nuclei are
densely packed together, which might cause commonly used bounding-box based
detection methods (and also pixel grouping methods) to struggle.
To address this problem,  we -- together with collaborators -- introduced a deep
learning based object detection and segmentation approach called
\stardist~\cite{schmidt2018,weigert2020}.
Instead of bounding boxes, \stardist represents objects with star-convex polygons, which are well suited for roundish objects such as cell nuclei.
Being primarily developed for fluorescence microscopy, we here aim to investigate \emph{i)} how \stardist can be extended to additionally perform classification of detected objects, \emph{ii)} whether it can be successfully applied to the different domain of histopathology images, and \emph{iii)} how it compares quantitatively against other methods used in histopathology (\eg HoverNet~\cite{graham2019short}) as part of the \emph{Colon Nuclei Identification and Counting (CoNIC)} challenge~\cite{conicshort}.

\begin{figure}[t]
  \centering
  \footnotesize
  \setlength{\tabcolsep}{1pt}
  \renewcommand{\arraystretch}{.5}
  \begin{tabular}{ccc}
    Input image & Ground truth & \stardistfns prediction\\[1pt]
    \includegraphics[width=0.32\linewidth]{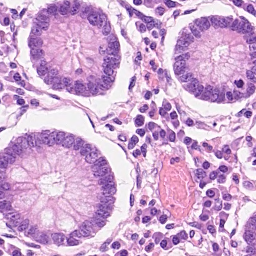} &
    \includegraphics[width=0.32\linewidth]{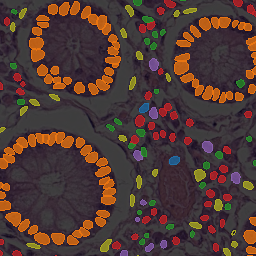} &
    \includegraphics[width=0.32\linewidth]{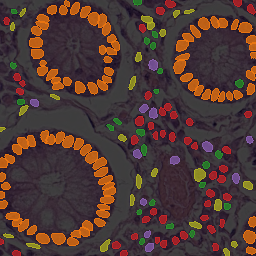}\\
    \includegraphics[width=0.32\linewidth]{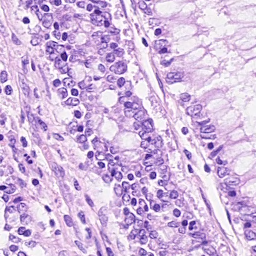} &
    \includegraphics[width=0.32\linewidth]{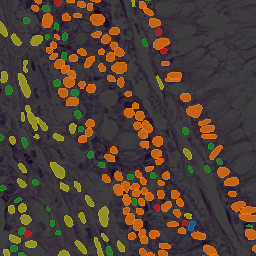} &
    \includegraphics[width=0.32\linewidth]{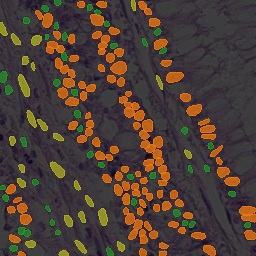}\\
    \includegraphics[width=0.32\linewidth]{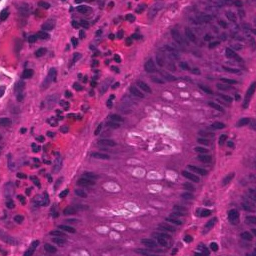} &
    \includegraphics[width=0.32\linewidth]{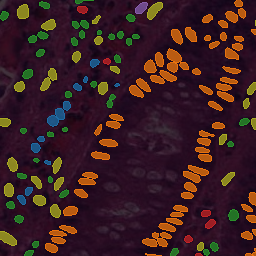} &
    \includegraphics[width=0.32\linewidth]{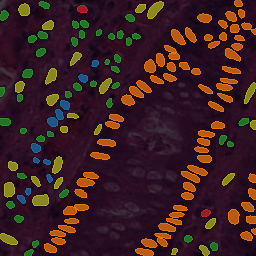}\\
    & \multicolumn{2}{c}{\includegraphics[width=.6\linewidth]{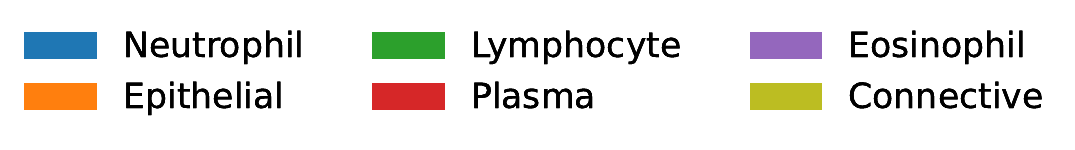}}%
  \end{tabular}%
  \vspace{-11pt}
  \caption{%
    Example input images from the \emph{CoNIC} challenge dataset (left)
    with ground truth (middle) and \stardist predictions (right). Each nucleus
    instance is shown with a color that reflects its class (see legend at
    bottom).
  }
  \vspace{-8pt}
  \label{fig:images}
\end{figure}

In this paper, we describe our extension of \stardist to perform nuclei classification and discuss several important adjustments that we made for our submissions to the \emph{CoNIC} challenge.
We evaluate key parameters via ablation experiments on the public challenge data
and by reviewing the results of our challenge submissions on the hidden preliminary and final test data.
First, we find that besides typical geometric data augmentations, special color augmentations to address staining variabilities in
\emph{hematoxylin} (H) and \emph{eosin} (E) are crucial to train models that generalize
well to new data.
Second, addressing the issue of imbalanced (cell) class distributions is
decisive to achieve high scores for the metrics used by the \emph{CoNIC} challenge, which
assign equal importance to each cell type.
Third, test-time augmentations and model ensembles do indeed substantially
boost performance.
Fourth, we find that a rather simple shape refinement procedure can additionally improve the
segmentation quality of \stardist.

\section{Method}
\label{sec:method}

\subsection{Extending \stardist for nuclei classification}
\label{sec:method:detection}
We shortly review the \stardist detection/segmentation approach as described in~\cite{schmidt2018,weigert2020}:
First, a convolutional neural network (CNN) takes a single input image and for
each pixel predicts \emph{1)} an \emph{object probability} to know if it is part
of an object, and \emph{2)} \emph{radial distances} to the boundary of the
object at that location (\ie~a star-convex polygon representation). %
Second, each pixel with an object probability above a chosen threshold votes for
a polygon to represent the shape of the object it belongs to. Since a given
object is potentially represented by many pixels which voted for its shape, a
non-maximum suppression (NMS) step is performed to prune the redundant polygons
that likely represent the same object.

The original \stardist~\cite{schmidt2018,weigert2020} approach only
allows for the prediction of individual shapes of object instances. To
additionally perform \emph{object instance classification}, we extend \stardist
by adding a \emph{semantic segmentation} head to the CNN backbone (besides
the existing two heads for object probabilities and radial distances). Now,
the CNN additionally predicts \emph{3)} \emph{class probabilities} for every
pixel.
After performing the object instance segmentation as described before, the class
(\ie, cell type)
of each object instance is determined by aggregating the class probabilities
from all pixels that are part of the respective instance.

Please note that all of our improvements and extensions of \stardist (as
compared to \cite{schmidt2018,weigert2020}) are included in our public code
repository.%
\footnote{We used the branch \texttt{\scriptsize
\href{https://github.com/stardist/stardist/tree/conic-2022}{\texttt{conic-2022}}} for all our
challenge submissions.}
Additional code developed to create our challenge submissions will be made
available after the challenge has ended.

\subsection{Data}

We only use the \emph{Lizard} dataset~\cite{lizardshort} for training our models,
specifically the extracted patches provided by the \emph{CoNIC} challenge organizers. The dataset consists of $4981$ images of size $256 \times 256 \times 3$ and corresponding label masks for the nuclei of six different cell types/classes: \emph{neutrophil, epithelial, lymphocyte, plasma, eosinophil}, and cells belonging to \emph{connective} tissue.
The distribution of cell classes is highly imbalanced: whereas epithelia nuclei constitute more than $60$\% of all objects, neutrophil and eosinophil nuclei each represent less than $1$\% of all instances.
Of all provided images we use $90$\% for training and $10$\% as (internal) validation set.
Besides dividing pixel values by $255$, we do not perform any image
preprocessing or data cleaning.

\subsection{Class balancing and augmentations}

We investigate different strategies to address the severe imbalance of the
distribution of nucleus types in the dataset. Concretely, we explore
\emph{1)} applying \emph{class weights} for the loss of the semantic segmentation head,
\emph{2)} using a \emph{focal loss}~\cite{lin2017} for the semantic segmentation head,
\emph{3)} and simple resampling/oversampling of the training data with a probability roughly proportional to the inverse of the class frequency (\ie, duplicating entire
image patches that contain many nuclei of the minority classes).
Somewhat surprisingly, we find that simple training data oversampling is by far the most effective strategy to combat class imbalance issues (\cref{sec:ablation}).

Based on our own augmentation library \emph{Augmend},%
\footnote{\texttt{\scriptsize \url{https://github.com/stardist/augmend}}
} %
we apply common geometric augmentations (flips and $90$ degree rotations,
elastic deformations) to each pair of input and label image \emph{on-the-fly}
during training.
Additionally, we explore different types of pixel-wise \emph{color
augmentations} by randomly changing \emph{1)} brightness, \emph{2)}
brightness and hue, or \emph{3)} brightness and H\&E staining.

\subsection{Model and Training details}

We use a \stardist model with $64$ rays and a U-Net~\cite{ronneberger2015}
backbone of depth $4$. For training, we apply a \emph{binary cross-entropy} loss for the
object probabilities, \emph{mean absolute error} for the radial distances, and
a sum of \emph{categorical cross-entropy} and \emph{Tversky loss}~\cite{abraham2019}
for the class probabilities.
Training was done from randomly initialized weights for $1000$ epochs ($256$
batches of size $4$) using the Adam~\cite{kingma2014} optimizer starting with a
learning rate of $0.0003$, which was reduced by half if no progress was made for
$80$ epochs. To reduce the effect of overfitting, we choose the weights for the final model that corresponded to the smallest validation loss during training.

\subsection{Test-time augmentations (TTA)}
\label{subs:tta}
Although data augmentation is heavily used during training, we find that
test-time augmentations (TTA) still lead to improved results.
To that end, we implement $8$ distinct geometric TTA and aggregate the
respective predictions to obtain more robust results.
Concretely, we apply all multiples of $90$ degree rotations (with and without
horizontal flips) to the input image and collect the CNN predictions for object
probabilities, radial distances, and class probabilities. Prediction tensors are
\emph{merged} by simple element-wise averaging.

Note that since the radial distances in \stardist refer to directions defined in polar
coordinates, these geometric transformations of the input image will result in a
permutation \wrt the order of the radial distances of the predicted tensor. For
example, a $90$ degree rotation of the input will cyclically shift the order of
the radial distances -- which has to be accounted for before merging the tensors
obtained via TTA.

\subsection{Postprocessing and shape refinement}
\label{subs:ppp}

As mentioned in \cref{sec:method:detection}, non-maximum suppression (NMS) is
applied as postprocessing after CNN prediction.
Concretely, NMS is performed here based on a set of polygon candidates (those
with object probability above a chosen threshold) obtained from the merged CNN
predictions (see above).
In each round of NMS, of the remaining polygons that haven't been suppressed,
the one with highest object probability is selected as the ``winner'' and will
suppress all other polygons that sufficiently overlap.
Instead of just keeping the winner polygon in each round to yield the final
object instances --~as is typically done in \stardist~-- we instead group each
winner polygon together with all the polygons that it suppressed.
For each group, we rasterize all polygons as binary masks
and aggregate them by majority vote
to obtain the mask of the given object instance.
We refer to this procedure as \emph{shape refinement}.

\subsection{Model ensembles}
We may additionally aggregate the predictions from a small number of separately
trained \stardist models. To that end, we collect the CNN predictions from each model
and merge them in the same way as described for TTA in \cref{subs:tta}.
This approach makes it also trivial to combine model ensembling with TTA, by simply
collecting and merging the TTA predictions from all models of the ensemble.
If it is desired to reduce the ensemble's computational requirements,
we can randomly sample only a few augmentations per model%
\footnote{We used $3$ or $4$ to stay within the time limit for the ensembles in \cref{tab:prediction}.}
(instead of using all $8$ possible geometric test-time augmentations).
Note that postprocessing and shape refinement (\cref{subs:ppp}) is only
performed once for the entire ensemble.

\section{Results}

We adopt the metrics%
\footnote{%
Note that the superscript $^+$ denotes that a metric is calculated over all
images of a dataset. Otherwise, the metric is computed separately for each image
of a dataset before the per-image values are averaged. } %
used by \emph{CoNIC} (see \cite{conicshort} for definitions) to evaluate the instance
segmentation and classification performance of our models. Concretely, the
overall performance is measured via the multi-class \emph{panoptic
quality}~\cite{kirillov2019}.
The panoptic quality $\PQ$ is defined as the product of \emph{detection quality} $\DQ$
($F_1$ score, \ie the harmonic mean of precision and recall) and
\emph{segmentation quality} $\SQ$ (average \emph{intersection over union} of all
correct matches).
The multi-class panoptic quality $\mPQ = \tfrac{1}{T} \sum_{t=1}^T \PQ_t$ %
is then defined as the average of the panoptic qualities $\PQ_t$
(only considering object instances of predicted class $t$)
for all $T$ cell classes/types.

\subsection{Ablation experiments}
\label{sec:ablation}
To investigate the effects of different \emph{a)} \emph{class balancing} approaches, \emph{b)} \emph{color augmentations}, and \emph{c)} \emph{test-time strategies}, we show the results of several ablation experiments%
\footnote{%
The default/baseline model uses oversampling for class balancing, brightness + H\&E staining color augmentations, and no test-time strategy.
}
on the \emph{internal} validation dataset in \cref{tab:ablation}.
Regarding class balancing, we find somewhat surprisingly that simple oversampling yields by far the best results for all metrics.
For color augmentations, our experiments seem to indicate that simple brightness augmentations are much more effective than augmentations that also affect hue or H\&E staining. This might be explained by the strong similarity between the \emph{internal} training and validation data. However, our observations are notably different regarding the results of our challenge submissions, where staining augmentations lead to considerably improved performance (\cf~\cref{tab:prediction}). This might be due to a more pronounced domain shift of the \emph{external} preliminary test data.
Finally, test-time augmentations and shape refinement lead to smaller improvements (relative to class balancing and color augmentations).

\begin{table}[t]
  \centering
  \small
\begin{tabular}{lcccc}
\toprule
 &             $\mPQ$ &              $\PQ$ &              $\DQ$ &              $\SQ$ \\
\midrule
\multicolumn{5}{l}{\emph{a) class balancing}}\\
             none &          0.3900 &          0.6841 &          0.4730 &          0.5510 \\
       focal loss &          0.4541 &          0.6711 &          0.5679 &          0.7882 \\
    class weights &          0.5099 &          0.6896 &          0.6281 &          0.8059 \\
     oversampling & \textbf{0.5885} & \textbf{0.6987} & \textbf{0.7186} & \textbf{0.8187} \\
\midrule
\multicolumn{5}{l}{\emph{b) color augmentation}}\\
       brightness & \textbf{0.6034} & \textbf{0.7037} & \textbf{0.7342} & \textbf{0.8218} \\
 brightness + hue &          0.5495 &          0.6850 &          0.6790 &          0.8074 \\
brightness + H\&E &          0.5884 &          0.6987 &          0.7186 &          0.8187 \\
\midrule
\multicolumn{5}{l}{\emph{c) test-time strategy}}\\
 shape refinement &          0.5832 &          0.6980 &          0.7053 &          0.8264 \\
              TTA &          0.5913 &          0.6980 &          0.7212 &          0.8192 \\
 TTA + shape ref. & \textbf{0.5984} & \textbf{0.7047} & \textbf{0.7225} & \textbf{0.8276} \\
             none &          0.5884 &          0.6987 &          0.7186 &          0.8187 \\
\bottomrule
\end{tabular} %
  \vspace{-4pt}
  \caption{Ablation results on the \emph{internal} validation set.}

  \label{tab:ablation}
\end{table}

\begin{table*}[t]
  \centering
  \small
  \addtolength\tabcolsep{-3pt}

\begin{tabular}{clrccccccccc}
  \toprule
\multicolumn{12}{c}{\emph{Preliminary test leaderboard (Task 1: segmentation and classification)}}\\
  \midrule
  Model &                      Strategy & Pos. &            $\mPQ^+$ &              $\PQ$ &             $\PQ^+$ &      $\PQ^+_{pla}$ &      $\PQ^+_{neu}$ &      $\PQ^+_{epi}$ &      $\PQ^+_{lym}$ &      $\PQ^+_{eos}$ &      $\PQ^+_{con}$ \\
  \midrule
    $A$ &                         basic                & 76  &          0.3647 &          0.5805 &          0.5822 &          0.3750 &          0.0877 &          0.5590 &          0.4391 &          0.3349 &          0.3929 \\
    $B$ &                      + H\&E staining augment & 30  &          0.4343 &          0.6380 &          0.6344 &          0.5029 &          0.1380 &          0.6265 &          0.4449 &          0.4652 &          0.4283 \\
    $\phantom{_2}B_2$ &                  + TTA         & 19  &          0.4515 &          0.6473 &          0.6452 &          0.5190 &          0.1639 &          0.6190 &          0.4614 &          0.5266 &          0.4188 \\
    $\phantom{_3}B_3$ &          + shape refinement    & 13  &          0.4583 &          0.6529 &          0.6512 &          0.5207 &          0.1812 &          0.6230 &          0.4711 &          0.5331 &          0.4205 \\
    $C$ &        + oversampling                        &  2  &          0.4970 &          0.6650 &          0.6625 &          0.5039 & \textbf{0.3407} &          0.6432 &          0.4748 &          0.5479 &          0.4715 \\
  $B,C,D$ & + ensemble                                   &  1  & \textbf{0.4971} & \textbf{0.6706} & \textbf{0.6669} & \textbf{0.5277} &          0.2819 & \textbf{0.6565} & \textbf{0.4887} & \textbf{0.5533} & \textbf{0.4745} \\
  \midrule
  \multicolumn{12}{c}{\emph{Final test leaderboard (Task 1: segmentation and classification)}}\\
  \midrule
  $C,D,E,F$ & as above &  1 &  0.5013 	&  0.6607  & 	0.6555 & -- & -- & -- & -- & -- & -- \\
\bottomrule
\end{tabular}   \vspace{-2pt}
    \caption{%
    \emph{Preliminary and final test leaderboard results} for submissions of our team {\small \texttt{EPFL\,|\,StarDist}} for the \emph{nuclear segmentation and classification} task of the \emph{CoNIC} challenge. \emph{Pos.}~denotes position at the end of the preliminary/final test phase.
    }

\label{tab:prediction}
\end{table*}

\subsection{CoNIC preliminary and final test phase submissions}

The \emph{CoNIC} challenge consists of two tasks: \emph{1) nuclear segmentation and classification} and \emph{2) prediction of cellular composition} (\ie, per-class nuclei counts).
As our aim is to improve \stardist, we focused on the first task and used the obtained results for task 2 by simply reporting the number of segmented nuclei per class.%
\footnote{%
While this seemingly did not perform very well during the preliminary test phase, it resulted in the third place on the final test leaderboard.
}
The \emph{preliminary test phase} allowed each team to make a limited number of submissions, which were quantitatively evaluated on a fraction of the final test set and the results shown publicly on a \emph{preliminary test leaderboard}.%
\footnote{\scriptsize \url{https://conic-challenge.grand-challenge.org/evaluation/challenge/leaderboard/}}
For the \emph{final test phase} only a single submission per team was allowed with the results on the complete test set being available on a \emph{final test leaderboard}.

\cref{tab:prediction} reports the results
of our team {\small\texttt{EPFL\,|\,StarDist}} %
for selected submissions on the preliminary test leaderboard as well for the final test leaderborard.
For the preliminary test phase, we first trained a basic \stardist model ($A$) with only standard geometric %
augmentations and class balancing via weighted loss terms. %
As we suspected a considerable domain shift in the test data, we next added heavy H\&E staining augmentations ($B$) that resulted in a large performance increase.
Without training a new model, we then added test-time augmentations ($B_2$) and shape refinement ($B_3$), each resulting in noticeable performance gains.
We finally realized that further addressing the class imbalance was likely to be the largest contributing factor to increase performance%
\footnote{While the challenge was still ongoing, the per-class metrics were also shown on the leaderboard -- but mixed up for some classes -- which suggested a very different cell type composition in the hidden test data. This was only noticed and communicated close to the end of the preliminary test phase.}
and thus aggressively oversampled the minority classes in the training data, leading to substantially improved results ($C$).
Finally, we similarly trained another model $D$ and submitted an ensemble of three models ($B$--$D$), yielding a slight improvement and resulting in the first place on the concluding preliminary test leaderboard.
For the final test set, we submitted an ensemble of four models ($C$--$F$), thereby combining two well performing models from the preliminary phase and two newly trained models $E$ and $F$, resulting in the top spot on the final test leaderboard of the CoNIC challenge.

\section{Discussion}

We described how \stardist can be successfully used and extended for object instance segmentation and classification in the context of histopathology.
Overall, our approach is competitive for the
segmentation and classification task of the \emph{CoNIC} challenge, which is demonstrated by winning the preliminary and final test phase.
 
\section{Compliance with ethical standards}
\label{sec:ethics}
No ethical approval was required for this study.

\section{Acknowledgments}
\label{sec:acknowledgments}

M.W.~is supported by the EPFL School of Life Sciences and a generous foundation represented by CARIGEST SA.

\bibliographystyle{IEEEbib}

\end{document}